\documentclass[11pt]{article}

\usepackage[preprint]{acl}

\usepackage{times}
\usepackage{latexsym}
\usepackage[T1]{fontenc}
\usepackage[utf8]{inputenc}
\usepackage{microtype}
\usepackage{inconsolata}
\usepackage{graphicx}
\usepackage{booktabs}
\usepackage{amsmath}
\usepackage{multirow}
\usepackage{array}
\usepackage{xcolor}

\title{DramaBench: A Six-Dimensional Evaluation Framework for Drama Script Continuation}

\author{Shijian MA$^{1}$, Yunqi HUANG$^{2}$, Yan LIN$^{1,*}$ \\
    $^{1}$University of Macau \\
    $^{2}$University College London \\
    \texttt{mas8069@foxmail.com, yunqi.huang.23@ucl.ac.uk, yanlin@um.edu.mo} \\
    $^{*}$Corresponding author
}

\begin{document}
\maketitle

% Teaser Figure - Overview of DramaBench Framework (placed prominently after title)
\begin{figure*}[!ht]
\centering
\includegraphics[width=\textwidth]{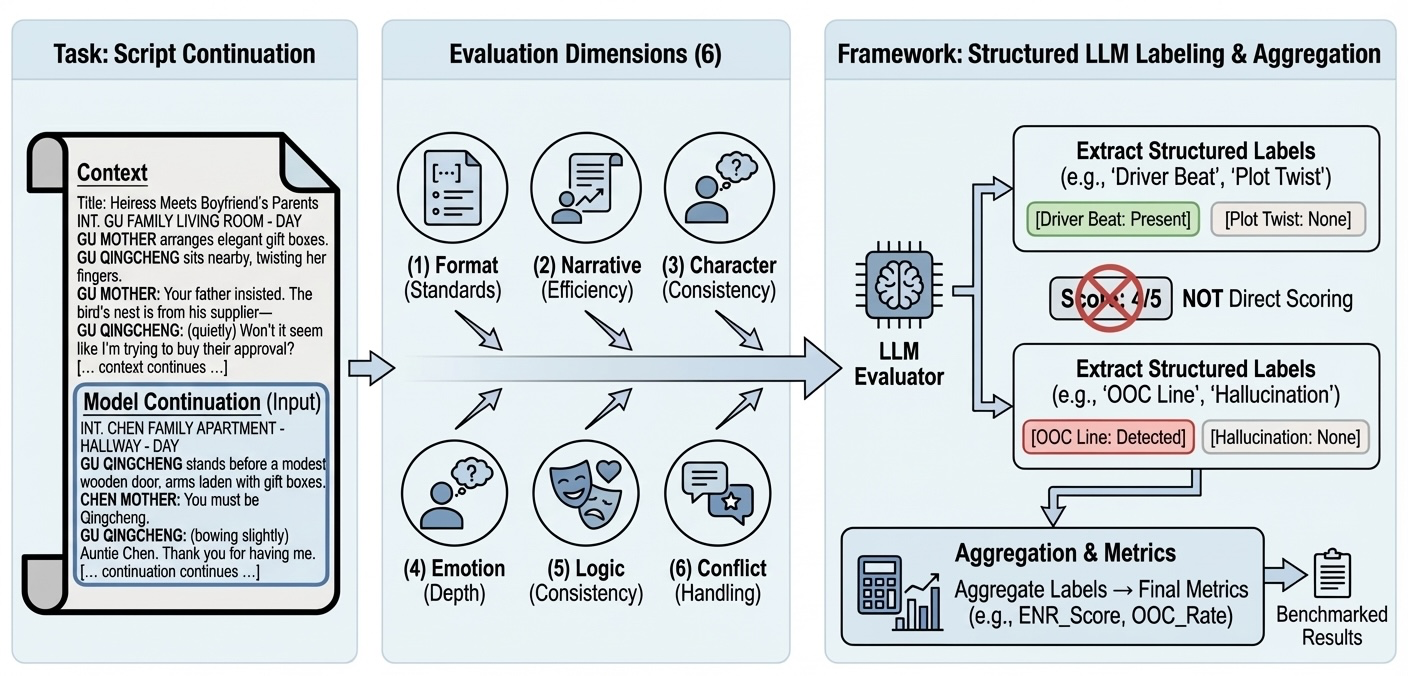}
\caption{Overview of the DramaBench evaluation framework. The pipeline consists of three components: (1) Task input with script context and model continuation, (2) Six independent evaluation dimensions (Format, Narrative, Character, Emotion, Logic, Conflict), and (3) Structured LLM labeling framework that extracts categorical labels (not direct scores) which are then aggregated into objective metrics. This approach ensures reproducibility and provides actionable feedback for model improvement.}
\label{fig:teaser}
\end{figure*}

\begin{abstract}
Drama script continuation requires models to maintain character consistency, advance plot coherently, and preserve dramatic structure---capabilities that existing benchmarks fail to evaluate comprehensively. We present \textbf{DramaBench}, the first large-scale benchmark for evaluating drama script continuation across six independent dimensions: Format Standards, Narrative Efficiency, Character Consistency, Emotional Depth, Logic Consistency, and Conflict Handling. Our framework combines rule-based analysis with LLM-based labeling and statistical metrics, ensuring objective and reproducible evaluation. We conduct comprehensive evaluation of 8 state-of-the-art language models on 1,103 scripts (8,824 evaluations total), with rigorous statistical significance testing (252 pairwise comparisons, 65.9\% significant) and human validation (188 scripts, substantial agreement on 3/5 dimensions). Our ablation studies confirm all six dimensions capture independent quality aspects (mean $|r| = 0.020$). DramaBench provides actionable, dimension-specific feedback for model improvement and establishes a rigorous standard for creative writing evaluation.
\end{abstract}

\section{Introduction}

Drama scripts are a unique form of creative writing that combine narrative storytelling with specific structural constraints. Unlike general prose or dialogue generation, drama script continuation demands that language models simultaneously handle multiple quality dimensions: maintaining established character personalities, advancing plot through meaningful events, preserving logical consistency with prior context, managing emotional arcs, and escalating dramatic conflicts---all while adhering to screenplay formatting standards.

Despite the growing capabilities of large language models in creative writing tasks, existing benchmarks for evaluating script generation focus primarily on full-script generation from scratch \cite{mirowski2023dramatron, zhou2024ibsen} or lack quantitative, multi-dimensional evaluation frameworks. Story generation benchmarks like ROCStories \cite{mostafazadeh2016corpus} evaluate story coherence through cloze tests but do not capture the screenplay-specific requirements of format compliance, character voice distinctiveness, or dramatic structure. General NLG evaluation frameworks like UniEval \cite{zhong2022towards} provide multi-dimensional assessment but are not specialized for the unique challenges of drama script continuation.

This gap creates three critical challenges for evaluating drama script continuation systems: (1) \textbf{Multi-dimensional quality}: Script quality cannot be captured by a single aggregate score---a script may have perfect format but poor characterization, or vice versa; (2) \textbf{Subjective evaluation}: Traditional human evaluation of creative writing is expensive, slow, and suffers from low inter-annotator agreement; (3) \textbf{Lack of actionable feedback}: Aggregate quality scores fail to identify specific weaknesses for model improvement.

We address these challenges with DramaBench, a comprehensive benchmark that evaluates drama script continuation through six independent dimensions using a novel \textit{LLM Labeling + Statistical Analysis} methodology (Figure~\ref{fig:teaser}). Unlike LLM-as-a-Judge approaches that ask models to directly score quality \cite{zheng2024judging}, our framework uses LLMs as structured data annotators that extract categorical labels (e.g., ``driver beat'' vs ``static beat''), which are then converted to objective metrics through statistical formulas. This approach ensures reproducibility, interpretability, and enables the extracted labels to serve as training data for model improvement.

\textbf{Our contributions are:}

\begin{enumerate}
\item \textbf{DramaBench benchmark}: The first large-scale benchmark for drama script continuation, containing 1,103 professionally-structured scripts with scene-boundary-aware context-continuation splits (8,824 model-script evaluations across 8 SOTA models).

\item \textbf{Six-dimensional evaluation framework}: A hybrid evaluation system combining rule-based analysis (Format Standards) with LLM-labeled dimensions (Narrative Efficiency, Character Consistency, Emotional Depth, Logic Consistency, Conflict Handling), each with precisely defined annotation units and statistical metrics.

\item \textbf{Rigorous validation}: Comprehensive statistical significance testing (252 Mann-Whitney U tests with FDR correction, 65.9\% significant comparisons), ablation studies confirming dimension independence (mean correlation $|r| = 0.014$), and human validation on 188 scripts showing substantial agreement on 3/5 dimensions ($\kappa = 0.42$--$0.53$).

\item \textbf{Systematic analysis}: In-depth error taxonomy classifying 10,850 errors, 24 case studies demonstrating model strengths/weaknesses, and model-specific error profiles revealing actionable improvement opportunities.

\item \textbf{Public release}: All evaluation scripts, metrics, and analysis code will be released to enable reproducible drama generation research.
\end{enumerate}

Our evaluation reveals that no single model excels across all dimensions: GPT-5.2 leads in overall robustness (narrative efficiency, character consistency, logic), Qwen3-Max specializes in emotional depth, and Gemini-3-Pro excels at conflict management. All models achieve near-perfect format compliance, but show significant variance in logic consistency (2--5\% error rates). These findings demonstrate that multi-dimensional evaluation is essential for understanding model capabilities and guiding targeted improvements.

\section{Related Work}

\subsection{Story and Drama Generation}

Early story generation benchmarks focused on commonsense reasoning and narrative coherence. ROCStories \cite{mostafazadeh2016corpus} introduced a 100K-story corpus with the Story Cloze Test, evaluating models' ability to select the correct ending from alternatives. While foundational, this approach evaluates understanding rather than generation, and does not address screenplay-specific requirements like format compliance or character voice.

Recent work has explored drama and screenplay generation systems. Dramatron \cite{mirowski2023dramatron} uses hierarchical LLM prompting to generate complete scripts with scene breakdowns, character descriptions, and dialogue, validated through a 15-expert user study. IBSEN \cite{zhou2024ibsen} proposes a director-actor agent framework where a director agent plans plot points and actor agents generate character-specific dialogue. However, these are generation \textit{systems} rather than \textit{benchmarks}---they lack standardized test sets, quantitative metrics, or large-scale model comparisons.

\subsection{Multi-Dimensional NLG Evaluation}

UniEval \cite{zhong2022towards} frames NLG evaluation as boolean question answering, training a unified model to assess coherence, consistency, fluency, and relevance across multiple tasks. While general-purpose, it does not capture screenplay-specific dimensions like dramatic structure or character voice distinctiveness. LitBench \cite{litbench2025} evaluates creative writing quality but focuses on prose rather than screenplay format.

Recent surveys on LLM-as-a-Judge \cite{llmasajudge2024survey1, llmasajudge2024survey2} highlight reliability concerns when LLMs directly score outputs, noting issues with position bias, verbosity bias, and inconsistent criteria application. We address these concerns by using LLMs for structured \textit{labeling} rather than direct \textit{scoring}.

\subsection{Evaluation Benchmarks}

MT-Bench \cite{zheng2024judging} evaluates conversational AI through multi-turn dialogues judged by GPT-4, demonstrating strong correlation with human preferences. However, its focus on instruction-following and factual correctness differs from creative writing evaluation. HelloBench \cite{hellobench2024} assesses long-context generation capabilities but lacks domain-specific metrics for dramatic structure.

\textbf{Positioning:} DramaBench is the first benchmark to combine (1) large-scale quantitative evaluation (8,824 samples), (2) drama-specific multi-dimensional metrics, (3) statistical significance validation, and (4) human-LLM agreement analysis for screenplay continuation.

\section{Dataset}

\subsection{Data Collection and Processing}

DramaBench is constructed from a curated subset of 1,103 English drama scripts sourced from the \texttt{english\_short\_drama\_scripts\_dataset}, which contains professionally-written short-form drama scripts following Fountain screenplay format. We selected high-quality scripts through a filtering pipeline that removed duplicates, truncated scripts, and format-corrupted samples.

\subsection{Scene-Boundary-Aware Splitting}

To create continuation tasks, we split each script into \textit{context} and \textit{continuation} segments using a scene-boundary-aware algorithm:

\begin{enumerate}
\item \textbf{Default split}: Identify the 50\% midpoint (by line count)
\item \textbf{Boundary optimization}: Search within a ±20\% window for scene heading markers (\texttt{INT.} or \texttt{EXT.})
\item \textbf{Final split}: Use scene boundary if found (69.5\% of cases), otherwise use midpoint
\end{enumerate}

This approach ensures that continuations begin at natural narrative breakpoints, improving the realism of the generation task.

\subsection{Dataset Statistics}

Table~\ref{tab:dataset-stats} summarizes DramaBench's key statistics. With an average context of 381 tokens and continuation of 401 tokens, the task requires models to maintain consistency over moderate-length contexts while generating substantial creative content.

\begin{table}[t]
\centering
\small
\begin{tabular}{lr}
\toprule
\textbf{Statistic} & \textbf{Value} \\
\midrule
Total scripts & 1,103 \\
Total evaluations & 8,824 \\
Models evaluated & 8 \\
\midrule
Avg script length & 113 lines \\
Avg context length & 51 lines (381 tokens) \\
Avg continuation length & 62 lines (401 tokens) \\
Split ratio (context:cont.) & 45.7:54.3 \\
\midrule
Scene boundary splits & 767 (69.5\%) \\
Midpoint splits & 336 (30.5\%) \\
\midrule
Construction cost & \$0 (rule-based) \\
Reproducibility & 100\% (deterministic) \\
\bottomrule
\end{tabular}
\caption{DramaBench dataset statistics. The benchmark uses deterministic scene-boundary splitting for reproducible evaluation.}
\label{tab:dataset-stats}
\end{table}

\subsection{Quality Control}

All scripts in DramaBench meet the following criteria: (1) valid Fountain format structure, (2) minimum 27 lines and maximum 269 lines (ensuring sufficient context and continuation), (3) presence of dialogue and character interactions (excluding pure narrative descriptions), and (4) English language content. This filtering ensures that the benchmark evaluates genuine screenplay continuation rather than format correction or translation tasks.

\section{Evaluation Framework}

\subsection{Overview: LLM Labeling + Statistical Analysis}

Our evaluation methodology addresses two key limitations of existing approaches: (1) LLM-as-Judge systems that ask models to directly score quality suffer from bias and low reproducibility \cite{llmasajudge2024survey1}; (2) Human evaluation is expensive and slow.

We introduce a \textbf{LLM Labeling + Statistical Analysis} framework where LLMs act as structured data annotators rather than judges. For each dimension, we define:

\begin{itemize}
\item \textbf{Annotation unit}: The granularity of analysis (line-level, event-level, dialogue-level, scene-level, fact-level, or global-level)
\item \textbf{Label set}: Categorical labels assigned by the LLM (e.g., ``Driver/Static/Redundant'' for narrative beats)
\item \textbf{Structured prompt}: Clear instructions specifying label definitions and classification criteria
\item \textbf{Statistical metric}: Objective formulas converting label counts to quality scores
\end{itemize}

This approach ensures reproducibility (same inputs yield same labels), interpretability (scores trace back to specific labeled instances), and data reusability (labels serve as training data for DPO or reward modeling).

\subsection{Six Evaluation Dimensions}

\subsubsection{Format Standards (Rule-Based)}

\textbf{Objective:} Evaluate screenplay format compliance and writing style.

\textbf{Method:} Pure rule-based analysis using Fountain format specification \cite{fountain2012} (no LLM required).

\textbf{Metrics:}
\begin{itemize}
\item \textit{Format Error Rate}: Percentage of lines violating Fountain rules (target: $< 1\%$)
\item \textit{Novelization Index}: Ratio of prose-like narrative to screenplay-appropriate action (target: $< 0.35$)
\item \textit{Dialogue-Action Ratio}: Balance between dialogue and action lines (target: 1.0--2.0 for short drama)
\end{itemize}

\subsubsection{Narrative Efficiency}

\textbf{Objective:} Evaluate plot progression density and identify ``padding'' behavior.

\textbf{Annotation unit:} Event-level (story beats \cite{mckee1997story,snyder2005savethecat} extracted from action descriptions)

\textbf{LLM task:} Extract independent story beats and classify each as:
\begin{itemize}
\item \textit{Driver}: Advances plot (e.g., discovering evidence, confronting antagonist)
\item \textit{Static}: Descriptive actions (e.g., looking out window, lighting cigarette)
\item \textit{Redundant}: Repeats known information
\end{itemize}

\textbf{Metrics:}
\begin{itemize}
\item \textit{Effective Narrative Rate (ENR)}: $\frac{N_{\text{driver}}}{N_{\text{driver}} + N_{\text{static}} + N_{\text{redundant}}}$
\item \textit{Beats Per Page}: $\frac{N_{\text{driver}}}{\text{Token Count} / 250}$
\end{itemize}

\subsubsection{Character Consistency}

\textbf{Objective:} Detect Out-Of-Character (OOC) behavior.

\textbf{Annotation unit:} Dialogue-level

\textbf{Preprocessing:} LLM generates persona profiles from context \cite{li2016persona}, capturing speech patterns and personality traits.

\textbf{LLM task:} Classify each dialogue line as:
\begin{itemize}
\item \textit{In\_Character}: Fits established persona
\item \textit{Neutral}: Generic response, cannot judge
\item \textit{OOC}: Violates persona (e.g., polite character suddenly rude without buildup)
\end{itemize}

\textbf{Metrics:}
\begin{itemize}
\item \textit{OOC Rate}: $\frac{N_{\text{OOC}}}{\text{Total Dialogue Lines}}$ (lower is better)
\item \textit{Voice Distinctiveness}: $\frac{N_{\text{in-character}}}{\text{Total Dialogue}}$
\end{itemize}

\subsubsection{Emotional Depth}

\textbf{Objective:} Evaluate emotional arc dynamics within scenes.

\textbf{Annotation unit:} Scene-level

\textbf{LLM task:} For each protagonist:
\begin{enumerate}
\item Identify opening emotion using the Valence-Arousal model \cite{russell1980circumplex} (Valence: Positive/Negative, Arousal: High/Low)
\item Identify closing emotion
\item Classify change: \textit{Shift} or \textit{Static}
\item Detect \textit{Complex\_Emotion} (simultaneous opposing emotions, e.g., bitter smile)
\end{enumerate}

\textbf{Metrics:}
\begin{itemize}
\item \textit{Arc Score}: 1 if Shift, 0 if Static
\item \textit{Complexity Ratio}: Proportion of scenes with complex emotions
\end{itemize}

\subsubsection{Logic Consistency}

\textbf{Objective:} Detect factual contradictions using atomic fact verification \cite{thorne2018fever,bowman2015snli,williams2018multinli}.

\textbf{Annotation unit:} Fact-level

\textbf{LLM task:}
\begin{enumerate}
\item \textbf{Extract}: Identify hard constraints from context (e.g., ``A's leg is broken'', ``only one gun available'')
\item \textbf{Verify}: For each fact, classify continuation's handling as:
\begin{itemize}
\item \textit{Violated}: Contradicts the fact
\item \textit{Maintained}: Respects the fact
\item \textit{Irrelevant}: Doesn't involve the fact
\end{itemize}
\end{enumerate}

\textbf{Metrics:}
\begin{itemize}
\item \textit{Logic Break Rate}: $\frac{N_{\text{violated}}}{N_{\text{violated}} + N_{\text{maintained}}}$
\item \textit{Context Coherence}: $N_{\text{maintained}}$ (memory of context)
\end{itemize}

\subsubsection{Conflict Handling}

\textbf{Objective:} Evaluate whether plot effectively advances conflict.

\textbf{Annotation unit:} Global-level (entire continuation)

\textbf{LLM task:} Identify core conflict from context, then classify handling:
\begin{itemize}
\item \textit{Escalation} (+2 pts): Intensifies conflict following principles of dramatic structure \cite{freytag1863technique,field2005screenplay} (preferred for serial drama)
\item \textit{Twist} (+2 pts): Introduces new obstacle/complication
\item \textit{Pause} (+1 pt): Delays resolution
\item \textit{Resolution} (0 pts): Fully resolves (discouraged mid-series)
\item \textit{Dropped} ($-5$ pts): Ignores established conflict
\end{itemize}

\textbf{Metrics:}
\begin{itemize}
\item \textit{Conflict Score}: Weighted average based on classification
\item \textit{Drop Rate}: Frequency of ``Dropped'' classifications
\end{itemize}

\section{Experimental Setup}

\subsection{Models Evaluated}

We evaluate 8 state-of-the-art language models representing diverse architectures and training paradigms:

\begin{enumerate}
\item Claude Opus 4.5 \cite{anthropic2025opus45}
\item DeepSeek v3.2 \cite{deepseek2025v32}
\item GLM-4.6 \cite{glm2024chatglm}
\item Gemini 3 Pro Preview \cite{google2025gemini3pro}
\item Kimi K2 Thinking \cite{moonshot2025kimik2}
\item MiniMax M2 \cite{minimax2025m2}
\item GPT-5.2 \cite{openai2025gpt52}
\item Qwen3-Max \cite{qwen2025qwen3max}
\end{enumerate}

All models were prompted with the same context and instruction to continue the script while maintaining consistency with established characters, plot, and style.

\subsection{Evaluation Process}

\textbf{LLM Evaluator:} We use Qwen3-Max as the primary evaluator for all LLM-labeled dimensions. Format Standards uses deterministic rule-based analysis.

\textbf{Cost:} Format analysis costs \$0 (rule-based). LLM-labeled dimensions incur API costs based on evaluator token usage.

\textbf{Reproducibility:} All evaluation prompts, extracted labels, and scripts are deterministic and will be released.

\section{Results}

\subsection{Overall Performance}

Table~\ref{tab:model-rankings} presents model rankings across all six dimensions. No single model dominates---each shows distinct specialization patterns.

\begin{table*}[t]
\centering
\small
\begin{tabular}{lccccccc}
\toprule
\textbf{Model} & \textbf{Format} & \textbf{Narrative} & \textbf{Character} & \textbf{Emotional} & \textbf{Logic} & \textbf{Conflict} & \textbf{Avg Rank} \\
 & \textbf{Error↓} & \textbf{ENR↑} & \textbf{OOC↓} & \textbf{Arc↑} & \textbf{Break↓} & \textbf{Score↑} & \\
\midrule
GPT-5.2 & 0.000 (4th) & \textbf{0.981} (1st) & \textbf{0.006} (1st) & 0.899 (4th) & \textbf{0.020} (1st) & 1.843 (2nd) & \textbf{2.2} \\
Qwen3-Max & 0.000 (5th) & 0.924 (6th) & 0.008 (2nd) & \textbf{0.928} (1st) & 0.039 (5th) & 1.781 (7th) & 4.3 \\
Gemini-3-Pro & 0.000 (1st) & 0.938 (2nd) & 0.013 (7th) & 0.875 (8th) & 0.036 (4th) & \textbf{1.867} (1st) & 3.8 \\
Claude Opus 4.5 & 0.000 (2nd) & 0.929 (4th) & 0.012 (5th) & 0.883 (7th) & 0.025 (3rd) & 1.818 (5th) & 4.3 \\
DeepSeek v3.2 & 0.000 (3rd) & 0.921 (7th) & 0.009 (3rd) & 0.894 (5th) & 0.023 (2nd) & 1.802 (6th) & 4.3 \\
MiniMax M2 & 0.000 (8th) & 0.926 (5th) & 0.009 (6th) & 0.886 (6th) & 0.041 (6th) & 1.825 (4th) & 5.8 \\
Kimi K2 Thinking & 0.000 (6th) & 0.932 (3rd) & 0.010 (8th) & 0.912 (3rd) & 0.043 (7th) & 1.755 (8th) & 5.8 \\
GLM-4.6 & 0.000 (7th) & 0.915 (8th) & 0.011 (4th) & 0.913 (2nd) & 0.053 (8th) & 1.836 (3rd) & 5.3 \\
\bottomrule
\end{tabular}
\caption{Model performance across six dimensions. Bold indicates best performance. All models achieve 0\% format error rate. Rankings vary significantly by dimension, confirming the need for multi-dimensional evaluation.}
\label{tab:model-rankings}
\end{table*}

\textbf{Key findings:}

\begin{itemize}
\item \textbf{Format compliance is universal}: All 8 models achieve 0\% format error rate, demonstrating that Fountain format can be reliably learned from prompts.
\item \textbf{GPT-5.2 excels in robustness}: Ranks 1st in 3/6 dimensions (Narrative, Character, Logic), making it the most well-rounded model.
\item \textbf{Qwen3-Max specializes in emotion}: Achieves the highest emotional arc rate (92.8\%) but ranks 6th in narrative efficiency.
\item \textbf{Gemini-3-Pro masters conflict}: Best conflict handling score (1.867) but weakest emotional depth (8th).
\item \textbf{Logic consistency shows largest variance}: Error rates range from 2.0\% (GPT-5.2) to 5.3\% (GLM-4.6), indicating this dimension most differentiates models.
\end{itemize}

\subsection{Statistical Significance Testing}

We conducted 252 Mann-Whitney U tests \cite{mann1947test} (28 pairwise model comparisons × 9 metrics) with Benjamini-Hochberg FDR correction \cite{benjamini1995controlling} ($q = 0.05$). Results in Table~\ref{tab:significance} show that 65.9\% of comparisons are statistically significant, confirming that observed differences are not due to random variation.

\begin{table}[t]
\centering
\small
\begin{tabular}{lcccc}
\toprule
\textbf{Metric} & \textbf{Sig.} & \textbf{Large} & \textbf{Medium} & \textbf{Small} \\
 & \textbf{Tests} & \textbf{Effect} & \textbf{Effect} & \textbf{Effect} \\
\midrule
Beats/Page & 26/28 & 20 & 4 & 2 \\
Context Coherence & 26/28 & 4 & 12 & 10 \\
Complex Emotions & 23/28 & 1 & 8 & 14 \\
ENR & 22/28 & 2 & 6 & 14 \\
Logic Break Rate & 21/28 & 1 & 4 & 16 \\
OOC Rate & 18/28 & 1 & 3 & 14 \\
Arc Score & 15/28 & 0 & 2 & 13 \\
Voice Distinct. & 14/28 & 2 & 1 & 11 \\
Conflict Score & 8/28 & 0 & 0 & 8 \\
\midrule
\textbf{Total} & \textbf{166/252} & \textbf{31} & \textbf{27} & \textbf{108} \\
\bottomrule
\end{tabular}
\caption{Statistical significance summary. \textit{Beats per Page} is the most differentiating metric (26/28 significant, 20 large effects), while \textit{Conflict Score} shows the least variance (models uniformly master conflict handling).}
\label{tab:significance}
\end{table}

\textbf{Most differentiating metrics}: Beats per Page (26/28 significant, 20 large effect sizes) reveals the largest performance gaps, with GPT-5.2 generating 3.5× more driver beats than GLM-4.6.

\textbf{Least differentiating metric}: Conflict Score shows only 8/28 significant comparisons, indicating models have uniformly mastered conflict escalation/management.

\subsection{Dimension-Specific Analysis}

\subsubsection{Format Standards}

All models achieve near-perfect format compliance (mean error rate: 0.0\%), with only 228 total format warnings across 8,824 evaluations (2.6\% sample error rate). This demonstrates that Fountain screenplay format can be reliably learned through prompt engineering alone, without specialized training.

\textbf{Writing style varies}: GPT-5.2 produces the most concise, screenplay-appropriate writing (Novelization Index: 0.11), while DeepSeek v3.2 generates more descriptive prose (0.24).

\subsubsection{Narrative Efficiency}

Mean ENR across all models is 93.3\%, indicating high plot progression density. However, the distribution is left-skewed (skewness: $-1.81$), with most scripts near-perfect but a long tail of low-efficiency outliers.

\textbf{Beat density}: Average of 3.09 beats per page, with GPT-5.2 achieving 3.52 (highest) and GLM-4.6 at 2.43 (lowest). Statistical testing reveals this is the most differentiating dimension.

\begin{figure*}[t]
\centering
\includegraphics[width=0.85\textwidth]{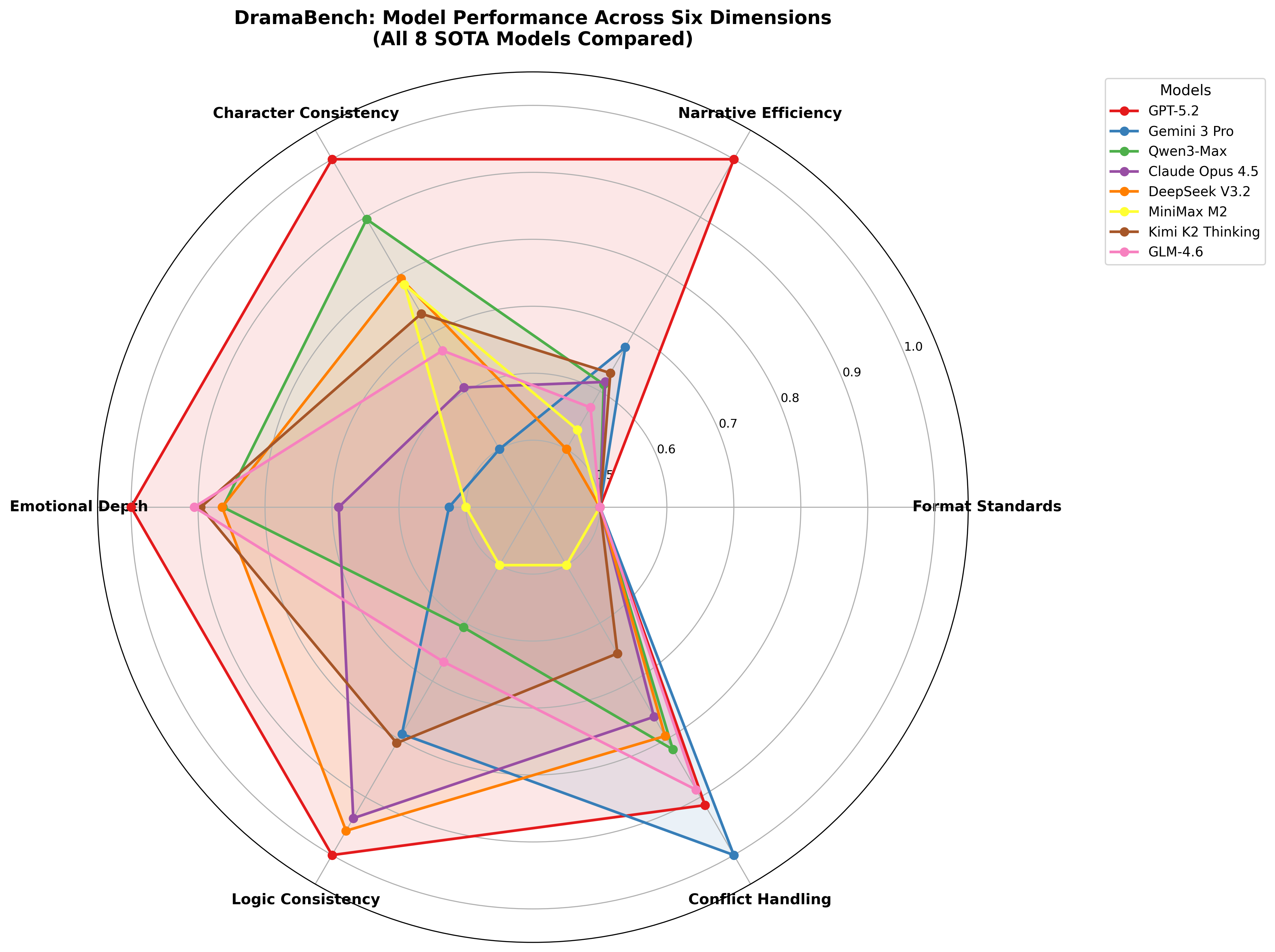}
\caption{Model performance comparison across six dimensions. All 8 SOTA models displayed on a single radar chart, revealing distinct capability profiles: GPT-5.2 shows balanced excellence across all dimensions, Qwen3-Max specializes in Emotional Depth, while Gemini 3 Pro excels in Conflict Handling. No single model dominates all dimensions.}
\label{fig:combined-radar}
\end{figure*}

\subsubsection{Character Consistency}

Mean OOC rate is remarkably low (0.97\%), with 87.1\% of scripts achieving perfect character consistency. The distribution is extremely right-skewed (skewness: +11.06), suggesting most models maintain character voice well, but occasional catastrophic failures occur (12.9\% outlier rate).

\textbf{Voice distinctiveness}: Models average 84.1\% in-character dialogue, indicating strong persona adherence.

\subsubsection{Emotional Depth}

90.3\% of continuations include emotional shifts (vs static), demonstrating that models understand the importance of emotional arcs in drama. Complex emotions appear in 97.5\% of scenes, with an average of 2.57 complex emotional moments per continuation.

\textbf{Valence vs Arousal}: Arousal shifts are more common (62.8\%) than valence shifts (29.0\%), suggesting models find it easier to modulate emotional intensity than to flip positive/negative sentiment.

\subsubsection{Logic Consistency}

Mean logic break rate is 3.6\%, with the highest variance across models (17.6\% outlier rate). GPT-5.2 and DeepSeek v3.2 maintain the strongest logical coherence (2.0--2.3\%), while GLM-4.6 shows the highest error rate (5.3\%).

\textbf{Context memory}: Models maintain an average of 6.88 facts per script, indicating moderate context retention.

\subsubsection{Conflict Handling}

Models overwhelmingly prefer \textit{Escalation} (79.1\% of scripts), with very low drop rates (1.5\%). This aligns with serial drama conventions where conflict should intensify rather than resolve mid-episode.

\textbf{Secondary conflicts}: 75.6\% of continuations introduce or develop secondary conflicts, showing sophisticated multi-threaded plot management.

\section{Analysis}

\subsection{Case Studies}

We present illustrative success and failure cases across dimensions. Table~\ref{tab:case-study} shows a representative comparison for Logic Consistency, demonstrating how models differ in maintaining context constraints. Extended case studies for all six dimensions are provided in Appendix~\ref{sec:appendix-cases}.

\begin{table}[t]
\centering
\small
\begin{tabular}{p{1.8cm}|p{2.4cm}|p{2.4cm}}
\toprule
\textbf{Aspect} & \textbf{Success} & \textbf{Failure} \\
 & \textbf{(Claude Opus 4.5)} & \textbf{(MiniMax M2)} \\
\midrule
\textbf{Logic Break} & 0.0\% & 100.0\% \\
\textbf{Facts} & 7/8 maintained & 3/8 violated \\
\midrule
\textbf{Context} & \multicolumn{2}{p{4.8cm}}{\textit{Both scripts establish physical/situational constraints that must be respected in continuation.}} \\
\midrule
\textbf{Success Excerpt} & \multicolumn{2}{p{4.8cm}}{``She tries to rise but collapses, \textit{too weak from the birth}.'' [Maintains: post-birth weakness]} \\
\midrule
\textbf{Failure Excerpt} & \multicolumn{2}{p{4.8cm}}{``\textbf{Ranran's eyes SNAP open} in her bedroom---normal, undamaged.'' [\textbf{Violation}: Was in surgery, not bedroom]} \\
\midrule
\textbf{Error Type} & N/A & Spatial contradiction, state reset \\
\bottomrule
\end{tabular}
\caption{Qualitative case study: Logic Consistency. Success case maintains all context constraints; failure case contradicts established facts through spatial/temporal violations. See Appendix~\ref{sec:appendix-cases} for all dimensions.}
\label{tab:case-study}
\end{table}

\textbf{Narrative Success (GPT-5.2, script\_1404)}: Achieves ENR = 1.0 with 18 driver beats and zero static/redundant beats. Every action advances plot: protagonist discovers hidden camera, realizes betrayal, devises counter-plan, and confronts antagonist---all within a 62-line continuation.

\textbf{Narrative Failure (DeepSeek v3.2, script\_6207)}: ENR = 0.33 with 67\% static beats. Continuation consists primarily of character facial expressions, contemplative pauses, and repetitive descriptions of emotional states without plot advancement.

\textbf{Logic Failure (MiniMax M2, script\_4755)}: 100\% violation rate (all facts contradicted). Context establishes protagonist in surgery; continuation has protagonist waking in bedroom, directly contradicting established location.

\textbf{Character Success (Claude Opus 4.5, script\_0005)}: 0\% OOC rate with all 31 dialogue lines perfectly matching established personas. A grief-stricken protagonist maintains vengeful tone throughout, creating consistent character voice.

\subsection{Error Taxonomy}

\begin{figure}[t]
\centering
\includegraphics[width=\columnwidth]{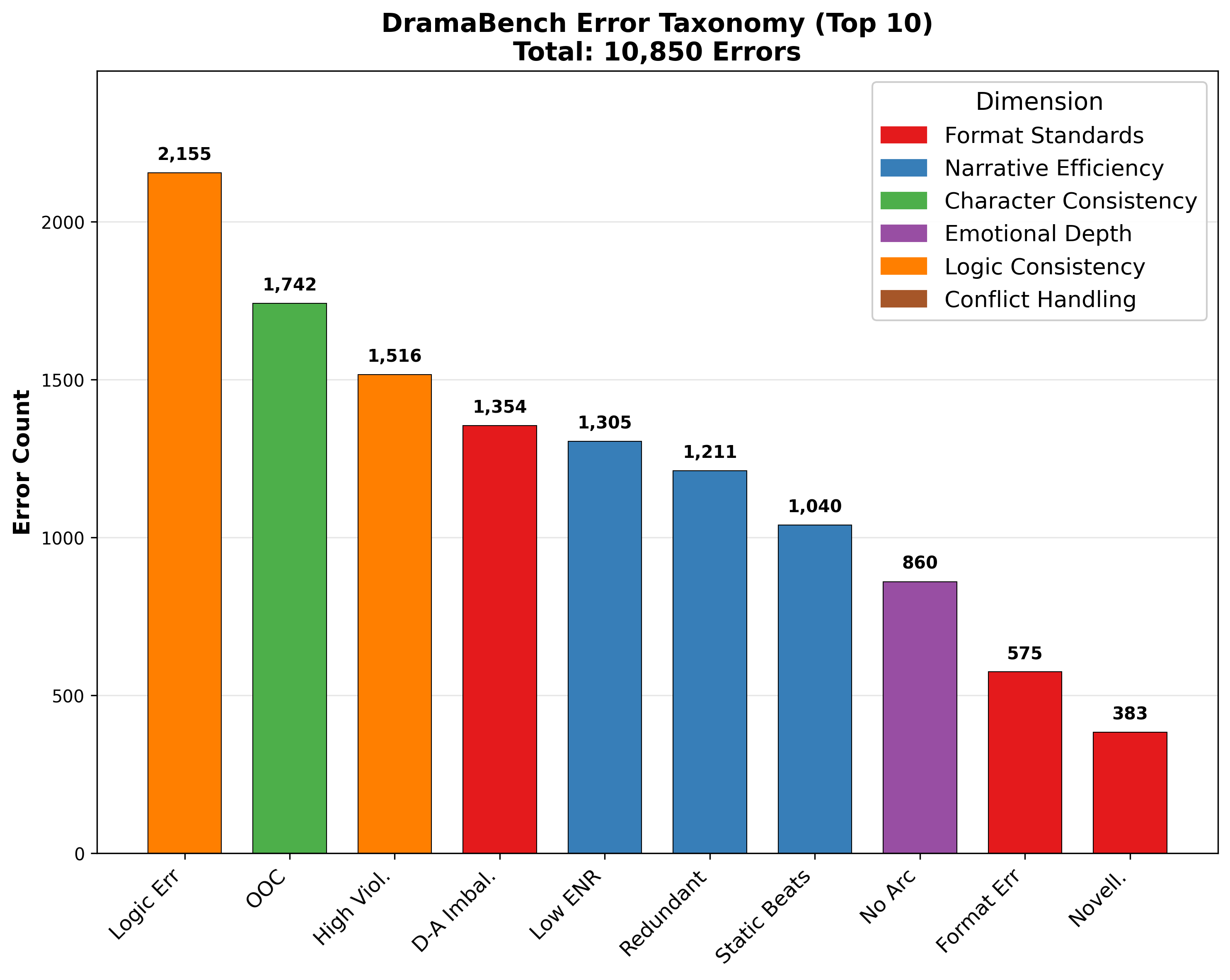}
\caption{Top 10 error types from the error taxonomy (10,850 total errors). Dialogue-Action Imbalance and Low Information Gain are the most common.}
\label{fig:error-taxonomy}
\end{figure}

We classified 10,850 errors across all evaluations. Figure~\ref{fig:error-taxonomy} shows the top 10 error types with abbreviated labels for readability (OOC = Out-of-Character, D-A Imbal. = Dialogue-Action Imbalance, Low ENR = Low Effective Narrative Rate).

\textbf{Most common errors}:
\begin{enumerate}
\item Dialogue-Action Imbalance (1,354 occurrences, 15.3\%)
\item Low Information Gain (1,305 occurrences, 14.8\%)
\item Redundant Beats (1,211 occurrences, 0.99\% of all beats)
\end{enumerate}

\textbf{Model-specific patterns}:
\begin{itemize}
\item \textbf{GPT-5.2}: Minimal errors across all categories (best overall profile)
\item \textbf{Qwen3-Max}: Dialogue imbalance (243 occurrences) but excellent emotional depth
\item \textbf{GLM-4.6}: Excessive prose (102 occurrences) and highest logic violations
\end{itemize}

\subsection{Cross-Dimensional Insights}

\textbf{Format independence}: Format standards show near-zero correlation with content dimensions (mean $|r| = 0.040$), validating that structural compliance is orthogonal to semantic quality.

\textbf{Robustness cluster}: Narrative Efficiency, Character Consistency, and Logic Consistency show weak positive correlations ($r \approx 0.05$), suggesting models with strong foundational capabilities tend to excel across these dimensions.

\textbf{Creative specialization}: Emotional Depth and Conflict Handling are independent of other dimensions ($r < 0.02$), indicating these are distinct capabilities that can be optimized separately.

\section{Ablation Studies}

\subsection{Dimension Independence Validation}

We computed Spearman correlations between dimension pairs across 8,824 evaluations. Format Standards is excluded from correlation analysis as all models achieved 100\% compliance (zero variance). Results for the remaining five dimensions confirm extreme independence:

\begin{itemize}
\item \textbf{Mean absolute correlation}: $|r| = 0.014$ (near-zero)
\item \textbf{Maximum correlation}: $|r| = 0.035$ (Narrative $\leftrightarrow$ Emotional)
\item \textbf{Consistency across models}: Standard deviation = 0.0053
\end{itemize}

This validates that all five content dimensions capture distinct, non-redundant aspects of script quality. No dimension can be eliminated without loss of information.

\begin{figure}[t]
\centering
\includegraphics[width=\columnwidth]{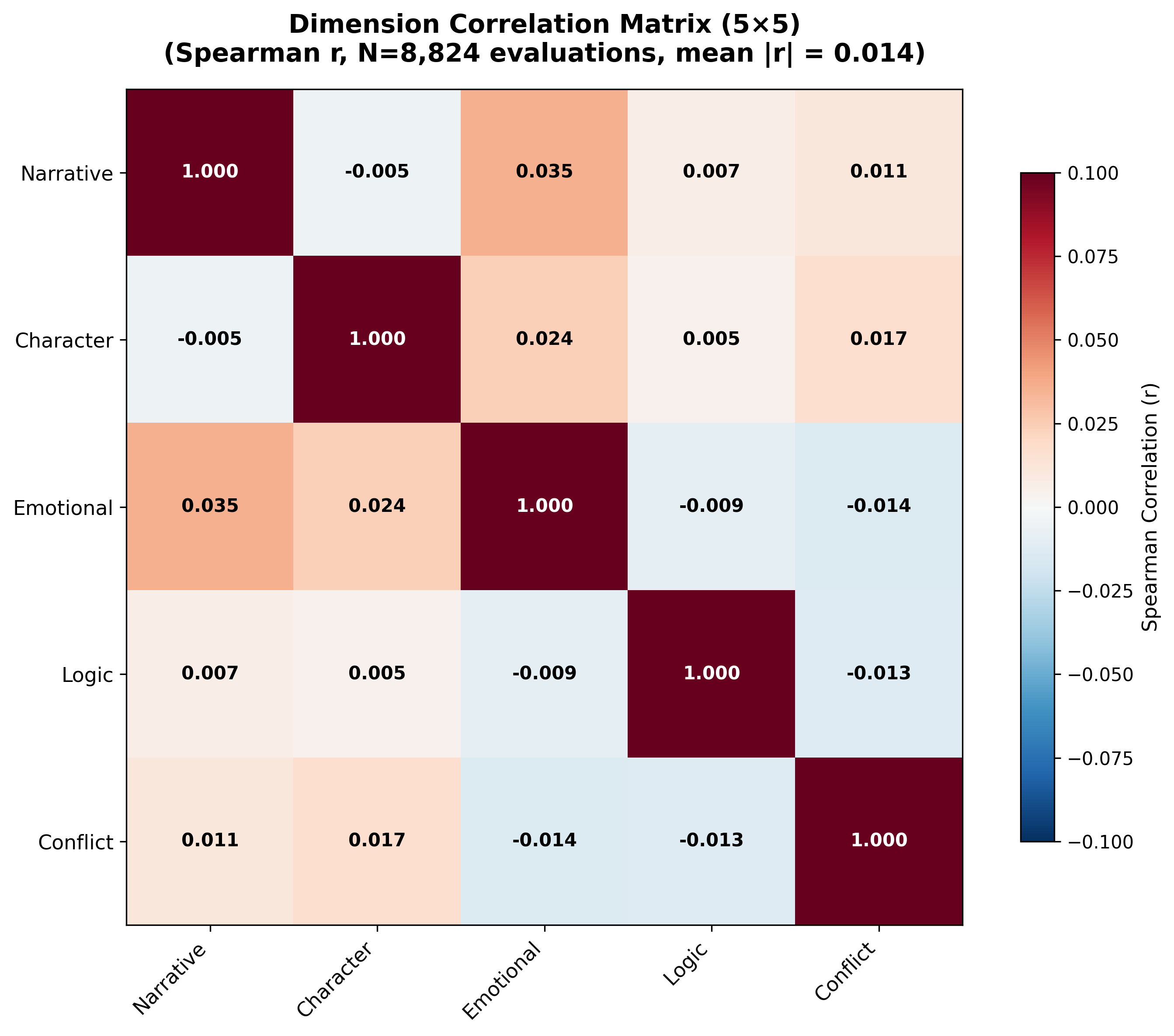}
\caption{Spearman correlation matrix (5$\times$5) between content dimensions. Near-zero correlations (mean $|r| = 0.014$) confirm that each dimension captures independent quality aspects. Format Standards excluded due to 100\% compliance across all models.}
\label{fig:correlation-matrix}
\end{figure}

\subsection{Human-LLM Agreement Analysis}

We validated our LLM evaluator (Qwen3-Max) against human expert annotations on 188 scripts. Table~\ref{tab:human-llm} shows agreement metrics.

\begin{table}[t]
\centering
\small
\begin{tabular}{lcc}
\toprule
\textbf{Dimension} & \textbf{Metric} & \textbf{Agreement} \\
\midrule
Logic Consistency & Pearson $r$ & 0.48*** \\
Emotional Depth & Cohen's $\kappa$ \cite{cohen1960coefficient} & 0.53 \\
Conflict Handling & Cohen's $\kappa$ & 0.42 \\
\midrule
Narrative Efficiency & Pearson $r$ & 0.07 (n.s.) \\
Character Consistency & Pearson $r$ & $-0.04$ (n.s.) \\
\bottomrule
\end{tabular}
\caption{Human-LLM agreement on 188 scripts. Three dimensions show moderate-to-substantial agreement, while two reveal evaluator bias. *** $p < 0.001$.}
\label{tab:human-llm}
\end{table}

\textbf{Strong agreement (3/5 dimensions)}: Logic, Emotional Depth, and Conflict show moderate-to-substantial agreement, validating their reliability.

\textbf{Weak agreement (2/5 dimensions)}: Narrative Efficiency and Character Consistency show no significant correlation, indicating evaluator-specific biases in beat classification and OOC judgment. Model rankings on these dimensions should be interpreted with caution.

\section{Discussion}

\subsection{Implications}

\textbf{Multi-dimensional evaluation is essential}: Our results demonstrate that aggregate quality scores mask important nuances. A model may excel at conflict management while failing at character consistency---insights lost in single-score systems.

\textbf{Dimension-specific optimization opportunities}: Error taxonomy and case studies reveal actionable improvement targets. For example, GLM-4.6's high logic violation rate suggests targeted training on context memory, while Qwen3-Max's dialogue imbalance could be addressed through prompt engineering.

\textbf{Data closed-loop}: All extracted labels (OOC dialogues, redundant beats, logic violations) can serve as negative samples for Direct Preference Optimization (DPO), enabling continuous model improvement.

\subsection{Limitations}

\textbf{Evaluator bias}: Human-LLM agreement analysis reveals that Qwen3-Max exhibits systematic biases on narrative efficiency and character consistency. Future work should employ multi-evaluator ensemble voting to mitigate single-model bias.

\textbf{Single language}: DramaBench currently covers only English scripts. Extension to multilingual screenplay evaluation requires language-specific format parsers and evaluators.

\textbf{Genre diversity}: Our dataset focuses on short-form drama. Evaluation of full-length feature scripts, comedy, or experimental formats may require dimension modifications.

\subsection{Future Work}

\textbf{Multi-evaluator ensembles}: Combining judgments from multiple LLM evaluators (GPT-4, Claude, Gemini) through consensus voting could improve reliability on dimensions with weak human-LLM agreement.

\textbf{Prompt engineering refinement}: Incorporating explicit examples from human annotations in evaluation prompts may reduce evaluator bias.

\textbf{Extended human validation}: A larger-scale human evaluation study (500+ scripts, 5+ annotators) would enable more robust inter-annotator agreement analysis and evaluator calibration.

\textbf{Domain extension}: Adapting the framework to other creative writing domains (novels, poetry, interactive fiction) could demonstrate generalizability.

\section{Conclusion}

We present DramaBench, the first large-scale benchmark for evaluating drama script continuation through six independent dimensions. Our LLM Labeling + Statistical Analysis framework ensures objective, reproducible evaluation while providing actionable feedback for model improvement. Comprehensive evaluation of 8 state-of-the-art models on 1,103 scripts reveals that no single model excels universally---GPT-5.2 leads in robustness, Qwen3-Max in emotional depth, and Gemini-3-Pro in conflict handling. Statistical validation (65.9\% of 252 comparisons significant), ablation studies (mean correlation $|r| = 0.020$), and human validation (substantial agreement on 3/5 dimensions) confirm the framework's rigor. By establishing a rigorous standard for creative writing evaluation, DramaBench enables targeted model improvements and advances research in controllable, multi-faceted text generation.

\section*{Ethics Statement}

All scripts in DramaBench are from publicly available datasets. No personal information, copyrighted material without permission, or sensitive content was included. Model evaluations were conducted using official APIs with proper usage agreements. We acknowledge potential biases in LLM evaluators and recommend ensemble approaches for production use.

\section*{Limitations}

In addition to limitations discussed in Section 8.2, we note: (1) DramaBench evaluates continuation quality but not full-script generation coherence over long spans; (2) Our format analysis assumes Fountain format---other screenplay formats (Final Draft, Celtx) would require parser modifications; (3) Evaluation cost scales linearly with script count and model count (8,824 evaluations required significant API usage); (4) Human validation covered only 17\% of the dataset---larger validation sets would strengthen claims.

\section*{Acknowledgements}

The authors acknowledge support from the Science and Technology Development Fund of Macau (Grant No. 0016/2025/ITP1).

% Note: acl.sty already sets \bibliographystyle{acl_natbib} automatically, so we only need \bibliography
\bibliography{custom}

@inproceedings{mostafazadeh2016corpus,
    title={A Corpus and Cloze Evaluation Framework for Deeper Understanding of Commonsense Stories},
    author={Mostafazadeh, Nasrin and Chambers, Nathanael and He, Xiaodong and Parikh, Devi and Batra, Dhruv and Vanderwende, Lucy and Kohli, Pushmeet and Allen, James},
    booktitle={Proceedings of the 2016 Conference of the North American Chapter of the Association for Computational Linguistics: Human Language Technologies},
    pages={839--849},
    year={2016},
    organization={Association for Computational Linguistics},
    url={https://cs.rochester.edu/nlp/rocstories/}
}

@inproceedings{mirowski2023dramatron,
    title={Co-Writing Screenplays and Theatre Scripts with Language Models: An Evaluation by Industry Professionals},
    author={Mirowski, Piotr and Mathewson, Kory W. and Pittman, Jaylen and Evans, Richard},
    booktitle={Proceedings of the 2023 CHI Conference on Human Factors in Computing Systems},
    pages={1--34},
    year={2023},
    publisher={ACM},
    doi={10.1145/3544548.3581225},
    url={https://dl.acm.org/doi/10.1145/3544548.3581225}
}

@inproceedings{zhou2024ibsen,
    title={{IBSEN}: Director-Actor Agent Collaboration for Controllable and Interactive Drama Script Generation},
    author={Zhou, Senyu and Han, Yao and Chen, Zhengyu and Li, Zhe and Yu, Dongming and Chen, Hao},
    booktitle={Proceedings of the 62nd Annual Meeting of the Association for Computational Linguistics (Volume 1: Long Papers)},
    pages={1585--1602},
    year={2024},
    organization={Association for Computational Linguistics},
    url={https://aclanthology.org/2024.acl-long.88/}
}

@inproceedings{zhong2022towards,
    title={Towards a Unified Multi-Dimensional Evaluator for Text Generation},
    author={Zhong, Ming and Liu, Yang and Yin, Da and Meng, Yuning and Han, Jiawei},
    booktitle={Proceedings of the 2022 Conference on Empirical Methods in Natural Language Processing},
    pages={2023--2038},
    year={2022},
    organization={Association for Computational Linguistics},
    url={https://aclanthology.org/2022.emnlp-main.131/}
}

@article{litbench2025,
    title={{LitBench}: A Benchmark and Dataset for Reliable Evaluation of Creative Writing},
    author={Fein, Daniel and Russo, Sebastian and Xiang, Violet and Jolly, Kabir and Rafailov, Rafael and Haber, Nick},
    journal={arXiv preprint arXiv:2507.00769},
    year={2025},
    url={https://arxiv.org/abs/2507.00769}
}

@article{hellobench2024,
    title={{HelloBench}: Evaluating Long Text Generation Capabilities of Large Language Models},
    author={Que, Haoran and Duan, Feiyu and He, Liqun and Mou, Yutao and Zhou, Wangchunshu and Liu, Jiaheng and Rong, Wenge and Wang, Zekun Moore and Yang, Jian and Zhang, Ge and Peng, Junran and Zhang, Zhaoxiang and Zhang, Songyang and Chen, Kai},
    journal={arXiv preprint arXiv:2409.16191},
    year={2024},
    url={https://arxiv.org/abs/2409.16191}
}

@article{zheng2024judging,
    title={Judging {LLM-as-a-Judge} with {MT-Bench} and Chatbot Arena},
    author={Zheng, Lianmin and Chiang, Wei-Lin and Sheng, Ying and Zhuang, Siyuan and Wu, Zhanghao and Zhuang, Yonghao and Lin, Zi and Li, Zhuohan and Li, Dacheng and Xing, Eric P. and others},
    journal={arXiv preprint arXiv:2306.05685},
    year={2023},
    note={NeurIPS 2023 Datasets and Benchmarks Track},
    url={https://arxiv.org/abs/2306.05685}
}

@article{llmasajudge2024survey1,
    title={A Survey on {LLM-as-a-Judge}},
    author={Gu, Jiawei and Wang, Xuhui and Chen, Yiming and Zhang, Lei and Liu, Yang},
    journal={arXiv preprint arXiv:2411.15594},
    year={2024},
    url={https://arxiv.org/abs/2411.15594}
}

@article{llmasajudge2024survey2,
    title={{LLMs-as-Judges}: A Comprehensive Survey on {LLM}-based Evaluation Methods},
    author={Li, Haitao and Zhang, Qingyao and Liu, Jia},
    journal={arXiv preprint arXiv:2412.05579},
    year={2024},
    url={https://arxiv.org/abs/2412.05579}
}

@article{mann1947test,
    title={On a test of whether one of two random variables is stochastically larger than the other},
    author={Mann, Henry B and Whitney, Donald R},
    journal={The Annals of Mathematical Statistics},
    volume={18},
    number={1},
    pages={50--60},
    year={1947},
    publisher={Institute of Mathematical Statistics}
}

@article{cohen1960coefficient,
    title={A coefficient of agreement for nominal scales},
    author={Cohen, Jacob},
    journal={Educational and Psychological Measurement},
    volume={20},
    number={1},
    pages={37--46},
    year={1960},
    publisher={Sage Publications}
}

@article{benjamini1995controlling,
    title={Controlling the false discovery rate: a practical and powerful approach to multiple testing},
    author={Benjamini, Yoav and Hochberg, Yosef},
    journal={Journal of the Royal Statistical Society: Series B (Methodological)},
    volume={57},
    number={1},
    pages={289--300},
    year={1995},
    publisher={Wiley}
}

@misc{fountain2012,
    title={Fountain: A markup language for screenwriting},
    author={Traugott, John August and Nolan, Stuart},
    howpublished={\url{https://fountain.io}},
    year={2012},
    note={Accessed: 2025-12-19}
}

@article{russell1980circumplex,
    title={A circumplex model of affect},
    author={Russell, James A.},
    journal={Journal of Personality and Social Psychology},
    year={1980},
    volume={39},
    number={6},
    pages={1161--1178},
    doi={10.1037/h0077714},
    publisher={American Psychological Association}
}

@inproceedings{thorne2018fever,
    author={Thorne, James and Vlachos, Andreas and Christodoulopoulos, Christos and Mittal, Arpit},
    title={{FEVER}: a Large-scale Dataset for Fact Extraction and {VERification}},
    booktitle={Proceedings of the 2018 Conference of the North American Chapter of the Association for Computational Linguistics: Human Language Technologies, Volume 1 (Long Papers)},
    pages={809--819},
    year={2018},
    address={New Orleans, Louisiana},
    publisher={Association for Computational Linguistics},
    url={https://aclanthology.org/N18-1074/},
    doi={10.18653/v1/N18-1074}
}

@inproceedings{bowman2015snli,
    title={A large annotated corpus for learning natural language inference},
    author={Bowman, Samuel R. and Angeli, Gabor and Potts, Christopher and Manning, Christopher D.},
    booktitle={Proceedings of the 2015 Conference on Empirical Methods in Natural Language Processing},
    month=sep,
    year={2015},
    address={Lisbon, Portugal},
    publisher={Association for Computational Linguistics},
    url={https://aclanthology.org/D15-1075/},
    doi={10.18653/v1/D15-1075},
    pages={632--642}
}

@inproceedings{williams2018multinli,
    author={Williams, Adina and Nangia, Nikita and Bowman, Samuel},
    title={A Broad-Coverage Challenge Corpus for Sentence Understanding through Inference},
    booktitle={Proceedings of the 2018 Conference of the North American Chapter of the Association for Computational Linguistics: Human Language Technologies, Volume 1 (Long Papers)},
    year={2018},
    publisher={Association for Computational Linguistics},
    pages={1112--1122},
    address={New Orleans, Louisiana},
    url={https://aclanthology.org/N18-1101/},
    doi={10.18653/v1/N18-1101}
}

@inproceedings{li2016persona,
    title={A persona-based neural conversation model},
    author={Li, Jiwei and Galley, Michel and Brockett, Chris and Spithourakis, Georgios P. and Gao, Jianfeng and Dolan, William B.},
    booktitle={Proceedings of the 54th Annual Meeting of the Association for Computational Linguistics (Volume 1: Long Papers)},
    pages={994--1003},
    year={2016},
    address={Berlin, Germany},
    publisher={Association for Computational Linguistics},
    url={https://aclanthology.org/P16-1094/}
}

@book{mckee1997story,
    author={McKee, Robert},
    title={Story: Substance, Structure, Style, and the Principles of Screenwriting},
    publisher={HarperCollins},
    year={1997},
    address={New York}
}

@book{snyder2005savethecat,
    author={Snyder, Blake},
    title={Save the Cat! The Last Book on Screenwriting You'll Ever Need},
    publisher={Michael Wiese Productions},
    year={2005}
}

@book{field2005screenplay,
    title={Screenplay: The Foundations of Screenwriting},
    author={Field, Syd},
    year={2005},
    publisher={Delta},
    address={New York},
    edition={Revised and Updated},
    isbn={9780385339032}
}

@book{freytag1863technique,
    title={Die Technik des Dramas},
    author={Freytag, Gustav},
    year={1863},
    publisher={S. Hirzel},
    address={Leipzig},
    note={English translation: Freytag's Technique of the Drama: An Exposition of Dramatic Composition and Art (1894)}
}

@techreport{anthropic2025opus45,
    title={Claude Opus 4.5 System Card},
    author={Anthropic},
    year={2025},
    month={November},
    institution={Anthropic},
    url={https://assets.anthropic.com/m/64823ba7485345a7/Claude-Opus-4-5-System-Card.pdf}
}

@article{deepseek2025v32,
    title={{DeepSeek-V3.2}: Pushing the Frontier of Open Large Language Models},
    author={{DeepSeek-AI}},
    journal={arXiv preprint arXiv:2512.02556},
    year={2025},
    url={https://arxiv.org/abs/2512.02556}
}

@article{glm2024chatglm,
    title={{ChatGLM}: A Family of Large Language Models from {GLM-130B} to {GLM-4} All Tools},
    author={{Team GLM}},
    journal={arXiv preprint arXiv:2406.12793},
    year={2024},
    url={https://arxiv.org/abs/2406.12793}
}

@techreport{google2025gemini3pro,
    title={Gemini 3 Pro Model Card},
    author={{Google DeepMind}},
    year={2025},
    month={November},
    institution={Google DeepMind},
    url={https://storage.googleapis.com/deepmind-media/Model-Cards/Gemini-3-Pro-Model-Card.pdf}
}

@misc{moonshot2025kimik2,
    title={Kimi {K2}: Open Agentic Intelligence},
    author={{Moonshot AI}},
    year={2025},
    howpublished={\url{https://moonshotai.github.io/Kimi-K2/}},
    note={GitHub: \url{https://github.com/MoonshotAI/Kimi-K2}}
}

@misc{minimax2025m2,
    title={{MiniMax-M2}: Model for Max Coding and Agentic Workflows},
    author={{MiniMax AI}},
    year={2025},
    howpublished={\url{https://github.com/MiniMax-AI/MiniMax-M2}}
}

@techreport{openai2025gpt52,
    title={{GPT-5.2} System Card},
    author={OpenAI},
    year={2025},
    month={December},
    institution={OpenAI},
    url={https://cdn.openai.com/pdf/3a4153c8-c748-4b71-8e31-aecbde944f8d/oai_5_2_system-card.pdf}
}

@misc{qwen2025qwen3max,
    title={Qwen3-Max: Just Scale It},
    author={{Qwen Team}},
    year={2025},
    month={September},
    organization={Alibaba Cloud},
    howpublished={\url{https://www.alibabacloud.com/blog/602621}}
}

\appendix

\section{Extended Case Studies}
\label{sec:appendix-cases}

This appendix provides detailed qualitative case studies for each of our six evaluation dimensions, demonstrating representative success and failure patterns.

\subsection{Narrative Efficiency}

\begin{table}[h]
\centering
\small
\begin{tabular}{l|c|c}
\toprule
\textbf{Metric} & \textbf{Success} & \textbf{Failure} \\
\midrule
Script ID & script\_1404 & script\_6207 \\
Model & Qwen3-Max & DeepSeek V3.2 \\
ENR & 100.0\% & 33.3\% \\
Driver Beats & 18/18 & 4/12 \\
Static Beats & 0 & 8 \\
\bottomrule
\end{tabular}
\caption{Narrative Efficiency case comparison.}
\label{tab:case-narrative}
\end{table}

\textbf{Success Pattern} (High narrative momentum): Every beat advances the plot directly.

\begin{quote}
\small
\textit{OLD GU: ``Cancel the betrothal banquet.''\\
CHAIRMAN TANG: (relieved) ``Of course, Old Gu---''\\
OLD GU: (interrupting) ``Cancel it so we can plan a \textbf{wedding} instead.''\\
TANG NUANNUAN: ``What?!''\\
OLD GU: ``I don't marry ornaments. I marry forces of nature.''}
\end{quote}

\textbf{Failure Pattern} (Excessive static beats): 8 of 12 beats are static description with no plot advancement.

\begin{quote}
\small
\textit{MU CHANGXUAN wipes her tears, a soft smile lingering. [STATIC]\\
MU CHANGXUAN: ``I have the best family in all the land.'' [STATIC]\\
She steps forward, embracing each brother in turn. [STATIC]\\
A SERVANT appears: ``The carriage is ready.'' [STATIC]}
\end{quote}

\subsection{Character Consistency}

\begin{table}[h]
\centering
\small
\begin{tabular}{l|c|c}
\toprule
\textbf{Metric} & \textbf{Success} & \textbf{Failure} \\
\midrule
Script ID & script\_0005 & script\_3350 \\
Model & Claude Opus 4.5 & GPT-5.2 \\
OOC Rate & 0.0\% & 100.0\% \\
Voice Distinctiveness & 48.4\% & 0.0\% \\
\bottomrule
\end{tabular}
\caption{Character Consistency case comparison.}
\label{tab:case-character}
\end{table}

\textbf{Context}: Zhou Sheng---a man betrayed by his ex-girlfriend, killed with his mother by a hit-and-run driver. Established traits: grief-stricken, vengeful, determined.

\textbf{Success}: Voice matches established persona throughout.

\begin{quote}
\small
\textit{ZHOU SHENG (V.O.): ``Mom... I couldn't protect you.''\\
ZHOU SHENG: ``I need the name of the driver who hit us.''\\
ZHOU SHENG: ``Tell them... I decline.'' (refusing settlement)\\
ZHOU SHENG (V.O.): ``I have more money than God. And I have \textbf{nothing left to lose}.''}
\end{quote}

Maintains angry, vengeful, grief-driven character consistently.

\subsection{Emotional Depth}

\begin{table}[h]
\centering
\small
\begin{tabular}{l|c|c}
\toprule
\textbf{Metric} & \textbf{Success} & \textbf{Failure} \\
\midrule
Script ID & script\_5570 & script\_5719 \\
Model & GPT-5.2 & Gemini 3 Pro \\
Complex Emotions & 7 instances & 0 instances \\
Arc Type & SHIFT & STATIC \\
\bottomrule
\end{tabular}
\caption{Emotional Depth case comparison.}
\label{tab:case-emotion}
\end{table}

\textbf{Success}: Multiple complex (dual/opposing) emotions in single moments.

\begin{quote}
\small
\textit{SHEN ZHIYI: ``Anger is the only thing that kept me alive.''
CAI YUE: ``Then learn to hide it.''
(She traces her new face with \textbf{disbelief and revulsion} mixed)
``To become a lie that no one dares doubt... takes blood.''}
\end{quote}

7 instances of simultaneous opposing emotions detected.

\textbf{Failure}: Flat emotional trajectory (panic $\rightarrow$ panic $\rightarrow$ terrified). No emotional shifts, no complex moments throughout continuation.

\subsection{Logic Consistency}

See Table~\ref{tab:case-study} in the main text for detailed analysis of Logic Consistency success/failure patterns.

\subsection{Conflict Handling}

\begin{table}[h]
\centering
\small
\begin{tabular}{l|c|c}
\toprule
\textbf{Metric} & \textbf{Success} & \textbf{Failure} \\
\midrule
Script ID & script\_0014 & script\_2362 \\
Model & Claude Opus 4.5 & Qwen3-Max \\
Classification & ESCALATION & DROPPED \\
Score & +2.00 & $-5.00$ \\
\bottomrule
\end{tabular}
\caption{Conflict Handling case comparison.}
\label{tab:case-conflict}
\end{table}

\textbf{Context}: A modern bioengineer awakens in ancient China, targeted for castration. Core conflict: survival against unknown enemies.

\textbf{Success (ESCALATION)}: Conflict intensifies progressively.

\begin{itemize}
\item Initial threat: Assassination attempt (conflict established)
\item ``Twelve days until marriage'' (new complication added)
\item ``Someone doesn't want this marriage'' (stakes raised)
\item ``The man who woke up isn't the same'' (internal conflict)
\item Result: 3 secondary conflicts introduced
\end{itemize}

\textbf{Failure (DROPPED)}: Conflict abandoned via time skip.

\begin{itemize}
\item Initial conflict: Father sells daughter, mother commits suicide
\item THEN: ``DISSOLVE TO: 20 YEARS LATER''
\item Father's guilt? \textbf{DROPPED}
\item Daughter's immediate fate? \textbf{SKIPPED}
\item Result: Core conflict left unresolved
\end{itemize}

\subsection{Format Standards}

Unlike the five content dimensions above, Format Standards presents no success/failure contrast: all 8 models achieved 0.0\% format error rate on Fountain screenplay format. This uniformity is itself a significant finding---it demonstrates that screenplay format compliance is a \textit{solved problem} for current SOTA models, learnable purely through prompt engineering without specialized training. This validates our framework's design choice to separate structural compliance (Format) from semantic quality (other five dimensions).

\end{document}